\documentclass[runningheads]{llncs}
\usepackage{graphicx}
\usepackage{textcomp}
\usepackage{tikz}
\usepackage{float}
\usepackage{caption}
\usepackage{multicol}
\usepackage{multirow}
\usepackage{tabularx}
\usepackage{booktabs}
\makeatletter
\newcommand{\printfnsymbol}[1]{%
  \textsuperscript{\@fnsymbol{#1}}%
}
\makeatother
\def\checkmark{\tikz\fill[scale=0.4](0,.35) -- (.25,0) -- (1,.7) -- (.25,.15) -- cycle;}

\begin{document}
\title{PNEL: Pointer Network based End-To-End Entity Linking over Knowledge Graphs}
%
%
 \author{Debayan Banerjee\inst{1} \and
 Debanjan Chaudhuri\thanks{equal contribution}\inst{2} \and \\
 Mohnish Dubey\printfnsymbol{1} \inst{2} \and
 Jens Lehmann\inst{2,3}}
 \authorrunning{D. Banerjee et al.}
%
 \institute{
  Language Technology Group, Universit{\"a}t Hamburg, Hamburg, Germany \email {banerjee@informatik.uni-hamburg.de}\and 
 Fraunhofer IAIS, Bonn/Dresden, Germany 
 \email{firstname.lastname@iais.fraunhofer.de}\and
Smart Data Analytics Group, University of Bonn \\
\email{jens.lehmann@cs.uni-bonn.de}
}
%

%
\maketitle              
\begin{abstract}
Question Answering systems are generally modelled as a pipeline consisting of a sequence of steps. In such a pipeline, Entity Linking (EL) is often the first step. Several EL models first perform span detection and then entity disambiguation. In such models errors from the span detection phase cascade to later steps and result in a drop of overall accuracy. Moreover, lack of gold entity spans in training data is a limiting factor for span detector training. Hence the movement towards end-to-end EL models began where no separate span detection step is involved. In this work we present a novel approach to end-to-end EL by applying the popular Pointer Network model, which achieves competitive performance. We demonstrate this in our evaluation over three datasets on the Wikidata Knowledge Graph.

\end{abstract}

\section{Introduction}
\begin{figure}[t]
\centering
\includegraphics[width=0.90\linewidth]{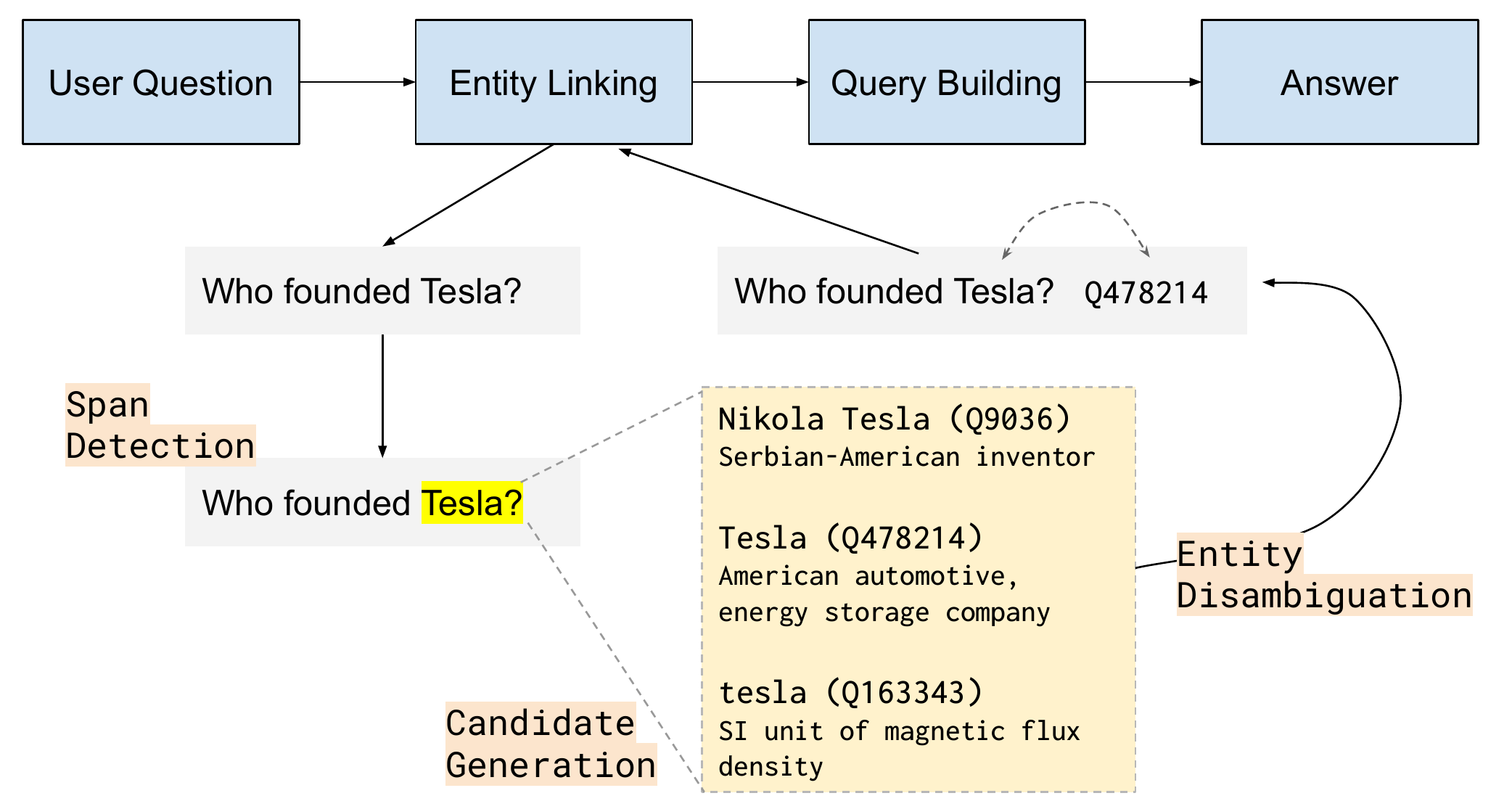}
\caption{Illustrating the use of Entity Linking in KGQA system.}
\label{pnel_exp}
\vspace{-10pt}
\end{figure}

Knowledge Graph based Question Answering (KGQA) systems use a background Knowledge Graph to answer queries posed by a user. Let us take the following question as an example (Figure \ref{pnel_exp}):  \textit{Who founded Tesla?}. The standard sequence of steps for a traditional Entity Linking system is as follows: The system tries to identify \textit{Tesla} as a span of interest. This task is called Mention Detection (MD) or Span Detection. Then an attempt is made to link it to the appropriate entity in the Knowledge Base. 
In this work we focus on Knowledge Bases in the form of graphs, hence the entity linker in this case tries to link \textit{Tesla} to the appropriate node in the graph. 
For a human, it is evident that the question is looking for a person's name who created an organisation named \textit{Tesla}, since the text contains the \textit{relation} \texttt{founded}. 
Hence, it is important that the entity linker understands the same nuance and ignores other entity nodes in the Knowledge Graph which also contain \textit{Tesla} in their labels, e.g., \texttt{Nikola Tesla (Q9036,
Serbian-American inventor), tesla (Q163343, SI unit)} when considering the example of the Wikidata knowledge graph. 
The task of ignoring the wrong candidate nodes, and identifying the right candidate node instead, is called \textit{Entity Disambiguation (ED)}. The cumulative process involving Mention Detection and Entity Disambiguation is called \textit{Entity Linking} (EL). 

Typically, the MD and ED stages are implemented by different machine learning models which require separate training. Especially for the MD part, sentences with marked entity spans are a requirement. In practice, such data is not easily available. Moreover, errors introduced by the MD phase cascade on to the ED phase. Hence, a movement towards end-to-end Entity Linkers began ~\cite{Kolitsas_2018} ~\cite{sorokin-gurevych-2018-mixing}. Such systems do not require labelled entity spans during training. In spite of the benefits of end-to-end models some challenges remain: Due to the lack of a span detector at the initial phase, each word of the sentence needs to be considered as an entity candidate for the disambiguation which leads to the generation of a much larger number of entity candidates. To re-rank these candidates a large amount of time is consumed, not just in processing the features of the candidates, but also in compiling their features. 

 In this work, we remain cognizant of these challenges and design a system that completely avoids querying the Knowledge Graph during runtime. PNEL (Pointer Network based Entity Linker) instead relies on pre-computed and pre-indexed TransE embeddings and pre-indexed entity label and description text as the only set of features for a given candidate entity. We demonstrate that this produces competitive performance while maintaining lower response times when compared to VCG ~\cite{sorokin-gurevych-2018-mixing}.

While there is a wide variety of KG embeddings to choose from, we confine our experiments to pre-computed TransE over Wikidata supplied by PyTorch-BigGraph~\cite{pbg}. Our choice was based on the popularity and ease of availability of these embeddings.

Traditionally, the Knowledge Graphs of choice for Question Answering research have been DBpedia~\cite{lehmann2015dbpedia}, Freebase ~\cite{bollacker2008freebase} and YAGO~\cite{suchanek2007semantic}. However, in recent times Wikidata~\cite{vrandevcic2014wikidata} has received significant attention owing to the fact that it covers a large number of entities (DBpedia 6M\footnote{https://wiki.dbpedia.org/develop/datasets/latest-core-dataset-releases}, Yago 10M\footnote{https://www.mpi-inf.mpg.de/departments/databases-and-information-systems/research/yago-naga/yago/}, Freebase 39M\footnote{\url{https://developers.google.com/freebase/guide/basic\_concepts\#topics}}, Wikidata 71M\footnote{\url{https://www.wikidata.org/wiki/Wikidata:Statistics}}). DBpedia, YAGO and Wikidata source their information from Wikipedia, however DBpedia and YAGO filter out a large percentage of the original entities, while Wikidata does not. While Wikidata has a larger number of entities it also adds to noise which is a challenge to any EL system. Wikidata also allows direct edits leading to up-to-date information, while DBpedia depends on edits performed on Wikipedia. Freebase has been discontinued and a portion of it is merged into Wikidata~\cite{10.1145/2872427.2874809}. Moreover DBpedia now extracts data directly from Wikidata, apart from Wikipedia \footnote{https://databus.dbpedia.org/dbpedia/wikidata}  ~\cite{10.1007/978-3-030-30796-7_7}.
Hence, we decide to base this work on the Wikidata knowledge graph and the datasets we evaluate on are all based on Wikidata.\\

In this work our \textbf{contributions} are as follows:

\begin{enumerate}

    \item PNEL is the first approach that uses the Pointer Network model for solving the End-to-End Entity Linking problem over Knowledge Graphs, inspired by the recent success of pointer networks for convex hull and generalised travelling salesman problems.
    \item We are the first work to present baseline results for the entire LC-QuAD 2.0~\cite{10.1007/978-3-030-30796-7_5} test set.
    \item Our approach produces state-of-the-art results on the LC-QuAD 2.0 and SimpleQuestions datasets.
\end{enumerate}

 The paper is organised into the following sections: (2) Related Work, outlining some of the major contributions in entity linking used in question answering; (3) PNEL, where we discuss the pointer networks and the architecture of PNEL
(4)Dataset used in the paper (5) Evaluation, with various evaluation criteria, results and ablation test  (6) Error Analysis (7) Discussion and future direction.

\section{Related Work}
\label{relatedwork}
DBpedia Spotlight~\cite{DBLP:conf/i-semantics/MendesJGB11} is one of the early works for entity linking over DBpedia. It first identifies a list of surface forms and then generates entity candidates. It then disambiguates the entity based on the surrounding context. In spite of being an early solution, it still remains one of the strongest candidates in our own evaluations, at the same time it has low response times. Compared to PNEL it lags behind in precision significantly.
S-MART~\cite{smart2015Yang}  generates multiple regression trees and then applies sophisticated structured prediction techniques to link entities to resources. S-MART performs especially well in recall on WebQSP in our evaluations and the reason seems to be that they perform more complex information extraction related tasks during entity linking, e.g., "Peter Parker" span fetches "Stan Lee" \footnote{https://github.com/UKPLab/starsem2018-entity-linking/issues/8\#issuecomment-566469263}. However compared to PNEL it has low precision. 

The journey towards end-to-end models which combine MD and ED in one model started with attempts to build feedback mechanisms from one step to the other so that  errors in one
stage can be recovered by the next stage. One of
the first attempts, Sil et al~\cite{10.1145/2505515.2505601}, use a popular NER model to generate extra number of spans and let
the linking step take the final decisions. Their method however depends on
a good mention spotter and the use of hand engineered features. It is also unclear how linking
can improve their MD phase. Later, Luo et al~\cite{luo-etal-2015-joint}
 developed competitive joint MD and ED models using semi-Conditional
Random Fields (semi-CRF). However, the basis for dependency was not robust, using only type-category correlation features. The other engineered features used in their model are either NER
or ED specific. Although their probabilistic
graphical model allows for low complexity learning and
inference, it suffers from high computational complexity caused by the usage of the cross product of all possible document spans, NER categories and entity assignments. Another solution J-NERD~\cite{nguyen-etal-2016-j} addresses the end-to-end task using engineered
features and a probabilistic graphical model on top
of sentence parse trees. EARL~\cite{10.1007/978-3-030-00671-6_7} makes some rudimentary attempts towards a feedback mechanism by allowing the entity and relation span detector to make a different choice based on classifier score in the later entity linking stage, however it is not an End-to-End model.

Sorokin et al~\cite{sorokin-gurevych-2018-mixing} is possibly the earliest work on end-to-end EL. They use features of variable granularities of context and achieve strong results on Wikidata that we are yet unable to surpass on WebQSP dataset. More recently, Kolitsas et al~\cite{Kolitsas_2018} worked on a truly end-to-end MD (Mention Detection) and ED (Entity Disambiguation) combined into a single EL (Entity Linking) model. They use context-aware mention embeddings, entity embeddings and a probabilistic mention - entity map, without demanding other engineered features.
Additionally, there are a few recent works on entity linking for short text on Wikidata~\cite{vrandevcic2014wikidata}, which is also the area of focus of PNEL. OpenTapioca~\cite{delpeuch2019opentapioca} works on a limited number of classes (humans, organisations and locations) when compared to PNEL, but is openly available both as a demo and as code and is lightweight. Falcon 2.0~\cite{sakor2019falcon} is a rule based EL solution on Wikidata which is openly available and fast, but it requires manual feature engineering for new datasets. 
Sevigli et al.~\cite{sevgili-etal-2019-improving} performs ED using KG entity embeddings (DeepWalk \cite{Perozzi_2014}) on Wikidata,  but they rely on an external MD solution. PNEL and Sorokin et al both use TransE entity embeddings and also perform MD and ED end-to-end in a single model.  Sorokin et al has a more complex architecture  when compared to PNEL. Apart from using TransE embeddings, they fetch neighbouring entities and relations on the fly during EL, which is a process PNEL intentionally avoids to maintain lower response times. KBPearl~\cite{10.14778/3384345.3384352} is a recent work on KG population which also targets entity linking as a task for Wikidata. It uses dense sub-graphs formed across the document text to link entities. It is not an end-to-end model but is the most recent work which presents elaborate evaluation on Wikidata based datasets, hence we include it in evaluations.\\
We also include QKBFly ~\cite{Nguyen:2017:QOK:3151113.3151119} and TagME ~\cite{10.1145/1871437.1871689} in our evaluations because KBPearl includes results for these systems on a common dataset (LC-QuAD 2.0). QKBFly performs on-the-fly knowledge base construction for ad-hoc text. It uses a semantic-graph representation of sentences that captures per-sentence clauses, noun phrases, pronouns, as well as their syntactic and semantic
dependencies. It retrieves relevant source documents for entity centric text from multiple souces like Wikipedia and other news websites. TagME is an older system that spots entity spans in short text using a Lucene index built out of anchor text in Wikipedia. It then performs a mutual-voting based disambiguation process among the candidates and finishes with a pruning step.

\section{PNEL}

\begin{figure*}[t]
\centering
\includegraphics[width=0.9\textwidth]{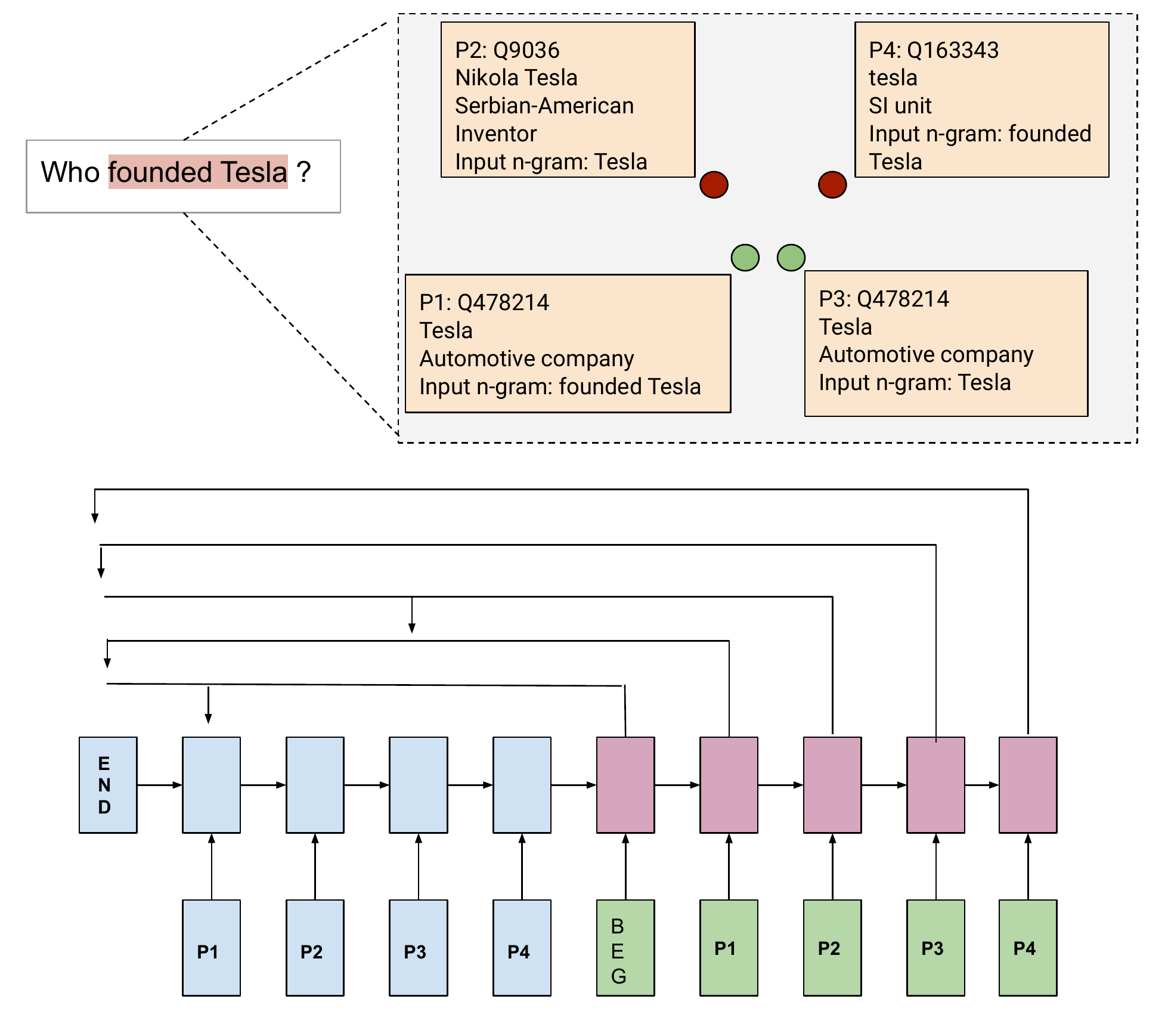}
\caption{The red and green dots represent entity candidate vectors for the given question. The green vectors are the correct entity vectors. Although they belong to the same entity they are not the same dots because they come from different n-grams. At each time step the Pointer Network points to one of the input candidate entities as the linked entity, or to the END symbol to indicate no choice. }
\label{pointer_network}
\end{figure*}

PNEL stands for Pointer Network based Entity Linker. Inspired by the use case of Pointer Networks \cite{NIPS2015_5866} in solving the convex hull and the generalised travelling salesman problems, this work adapts the approach to solving entity linking. \textit{Conceptually, each candidate entity is a point in an euclidean space, and the pointer network finds the correct set of points for the given problem}.

\subsection{Encoding for Input}

\begin{figure*}[!ht]
\centering
\includegraphics[width=0.75\linewidth]{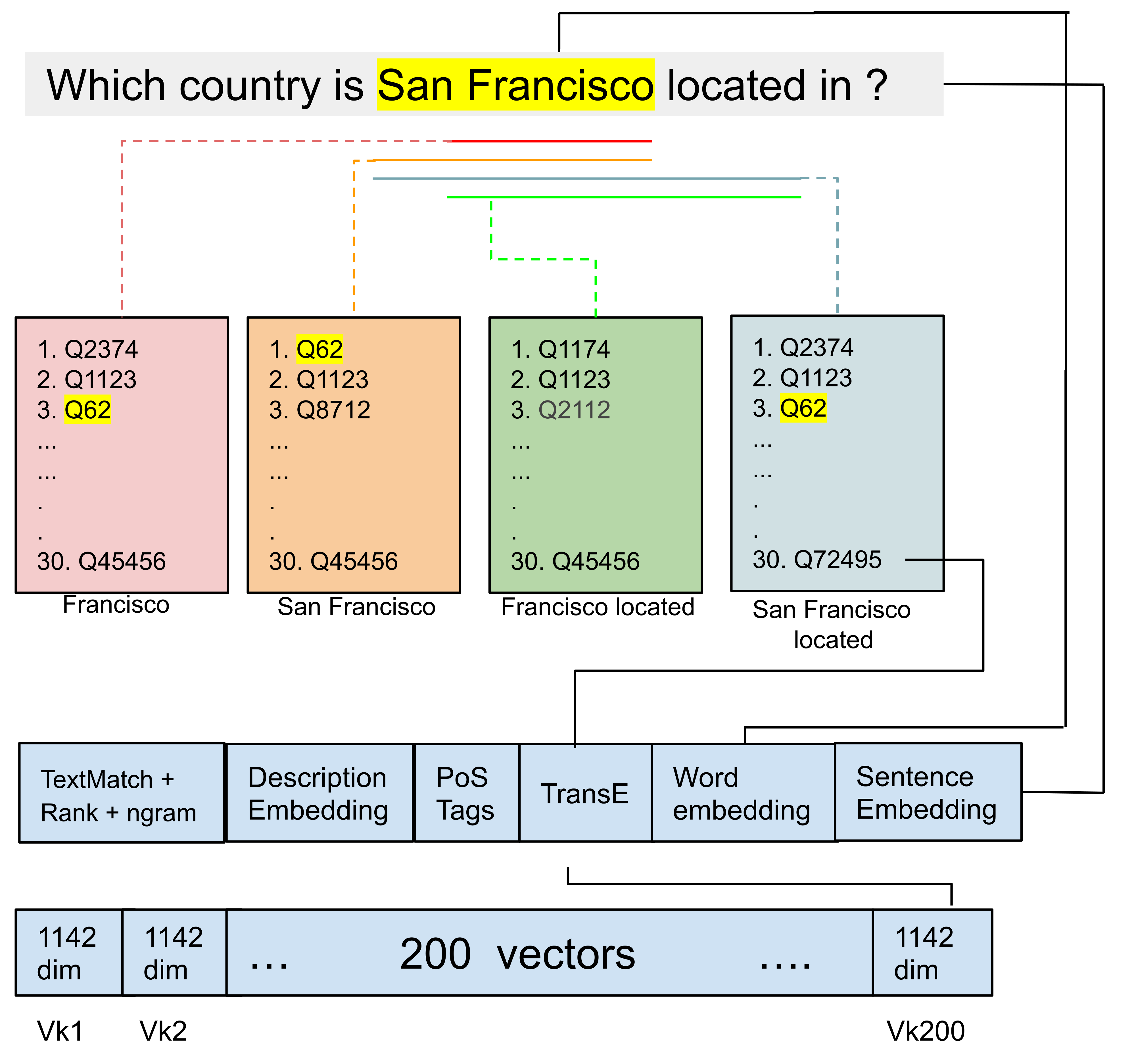}
\caption{The word ``Francisco" is vectorised in the following manner: 4 n-grams represented by the underlines are considered and searched against an entity label database. The top 50 search results are depicted for each of the n-grams resulting in 200 candidates. For the entity Q72495, for example, we fetch its TransE embedding, add its text search rank, n-gram length, word index position as features. Additionally we also append the fastText embedding for ``Francisco" and the entire fastText embedding for the sentence (average of word vectors) to the feature. We then append the fastText embeddings for the label and description for this entity. Hence we get a 1142 dimensional vector \textbf{$V_{k120}$ corresponding to entity candidate Q72495}. For all 200 candidate entities for ``Francisco", we have a sequence of two hundred 1142 dimensional vectors as input to the pointer network. For the sentence above which has 7 words, this results in a final sequence of $7\times200=1400$ vectors each of length 1142 as input to our pointer network. Any one or more of these vectors could be the correct entities.  }
\label{vectorisation}
\vspace{-10pt}
\end{figure*}

The first step is to take the input sentence and vectorise it for feeding into the pointer network. We take varying length of n-grams, also called n-gram tiling and vectorise each such n-gram.

Given an input sentence $S = \{s_1,s_2...s_n\}$ where $s_k$ is a token (word) in the given sentence, we vectorise $s_k$ to $v_k$, which is done in the following manner:

\begin{enumerate}
    \item Take the following 4 n-grams: $[s_k], [s_{k-1},s_k],[s_k,s_{k+1}],[s_{k-1},s_k,s_{k+1}]$
    \item For each such n-gram find the top $L $ text matches in the entity label database. We use the OKAPI BM25 algorithm for label search.
    \item For each such candidate form a candidate vector comprising of the concatenation of the following features
    \begin{enumerate}
        \item $R_{kl}$ = Rank of entity candidate in text search (length 1), where $ 1 \leq l \leq L$
        \item $ngramlen$ = The number of words in the current n-gram under consideration where $ 1 \leq ngramlen \leq 3$ (length 1)
        \item $k$ = The index of the token $s_k$ (length 1)
        \item $pos_k$ = A one-hot vector of length 36 denoting the PoS tag of the word under consideration. The 36 different tags are as declared in the Penn Treebank Project ~\cite{santorini1990part}.  (length 36)
        \item $EntEmbed_{kl}$ = TransE Entity Embedding (length 200)
        \item $SentFTEmbed$ = fastText embedding of sentence $S$ (length 300), which is a mean of the embeddings of the tokens of $S$. In some sense this carries within it the problem statement.
        \item $TokFTEmbed_{k}$ = fastText embedding of token $s_k$ (length 300). Addition of this feature might seem wasteful considering we have already added the sentence vector above, but as shown in the ablation experiment in Experiment \ref{pnelablation}, it results in an improvement.
        \item $DescriptionEmbed_{kl}$ = fastText embedding of the Wikidata description for entity candidate $kl$ (length 300)
        \item $TextMatchMetric_{kl}$ = This is a triple of values, each ranging from 0 to 100, that measures the degree of text match between the token under consideration $s_k$ and the label of the entity candidate $kl$. The three similarity matches are \textit{simple ratio, partial ratio, and token sort ratio}. In case of \textit{simple ratio} the following pair of text corresponds to perfect match: \texttt{"Elon Musk" and "Elon Musk"}. In case of \textit{partial ratio} the following pair of text corresponds to a perfect match: \texttt{"Elon Musk" and "Musk"}. In case of \textit{token sort ratio} the following pair of text corresponds to a perfect match: \texttt{"Elon Musk" and "Musk Elon"}. (length 3)
    \end{enumerate}
\end{enumerate}

For each token $s_k$ we have an expanded sequence of token vectors, comprising of 4 n-grams, upto 50 candidates per n-gram, where each vector is of length 1142. Hence each token $s_k$ is transformed into $4 \times 50  = 200$ vectors, each a 1142 length vector (see Figure \ref{vectorisation}). We may denote this transformation as $s_k \rightarrow \{v_{k1},v_{k2}....v_{k200}\}$. Note that there may be less than 50 search results for a given token so there may be less than 200 entity candidates in the final vectorisation. Each of these $v_k$ vectors is an entity candidate.

\begin{figure*}[thb]
  \centering
  \begin{minipage}[b]{0.49\textwidth}
    \includegraphics[width=1.0\textwidth]{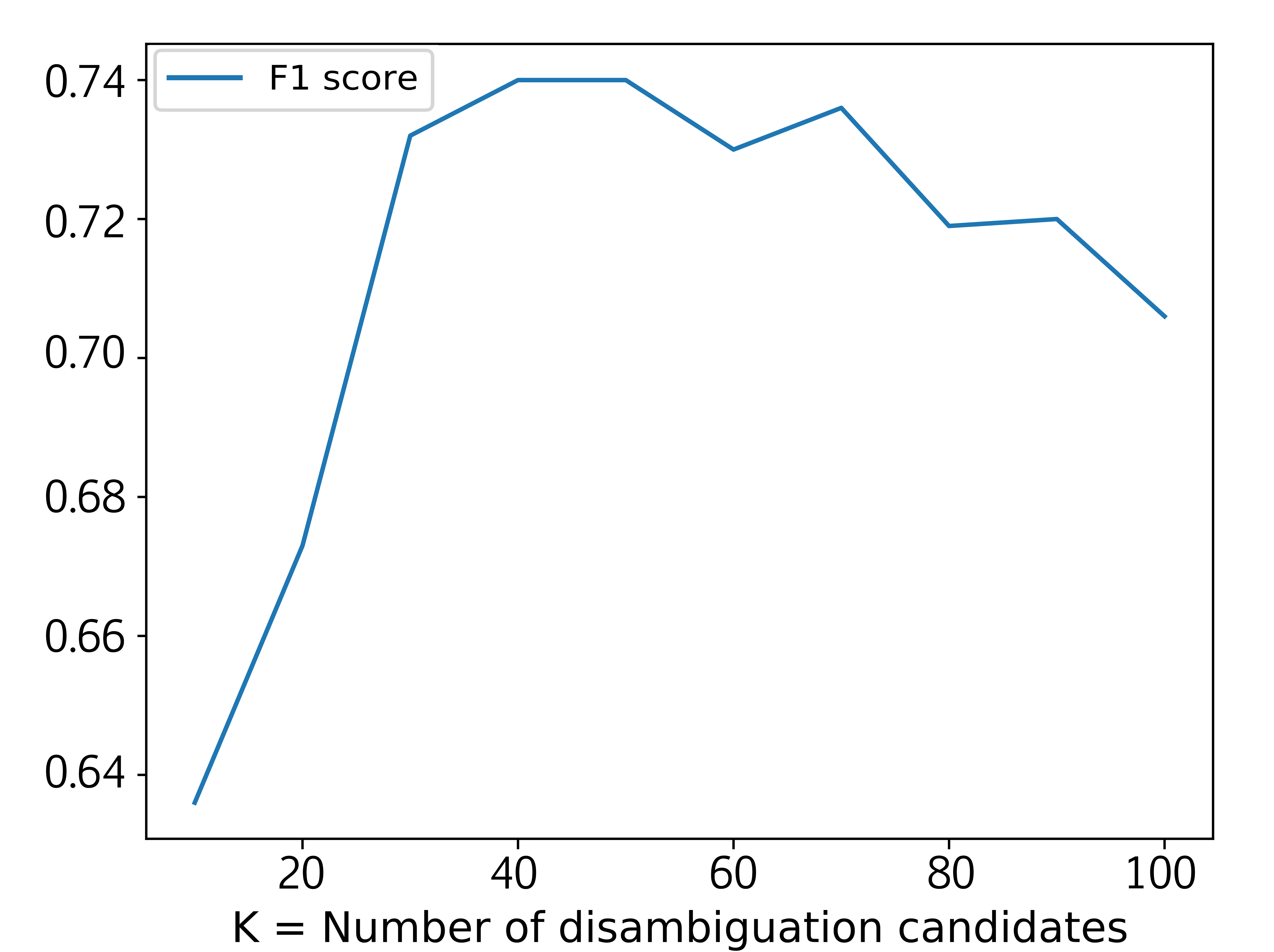}
  \end{minipage}
  \hfill
  \begin{minipage}[b]{0.49\textwidth}
    \includegraphics[width=1.0\textwidth]{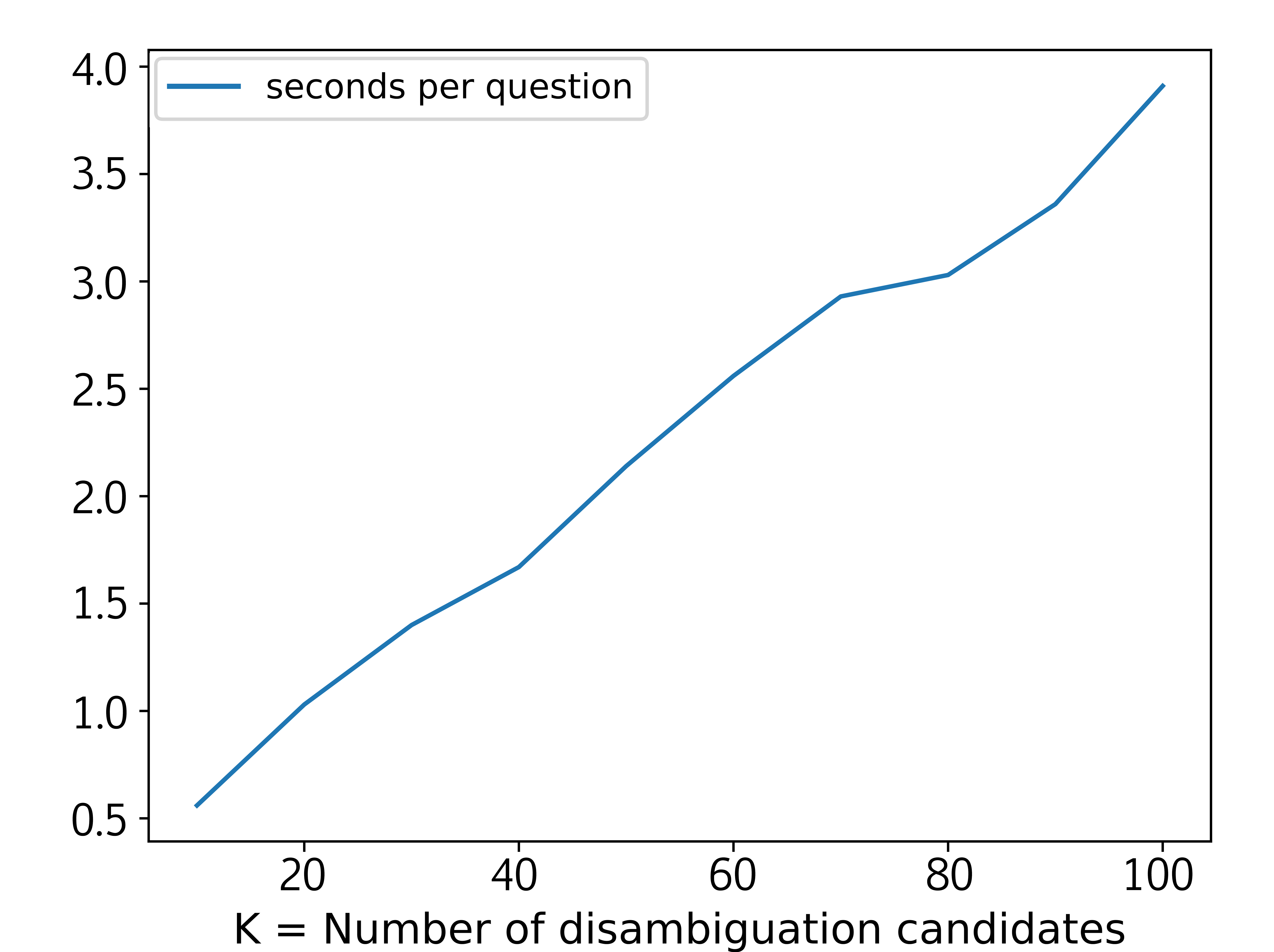}
  \end{minipage}
  \hfill
      \caption{K=Number of search candidates per n-gram. On the left: K vs F1 score on a set of 100 WebQSP test questions, with average word length of 6.68. F1 is maximum for K=40 and 50. On the right: K vs time taken for PNEL to return a response. The relationship appears to be close to linear.}
\label{kf1}
\end{figure*}

\subsection {Training}

For the entire sentence, a sequence of such vectors is provided as input to the pointer network. During training the labels for the given input sequence are the index numbers of the correct entities in the input sequence. Note that the same entity appears multiple times because of n-gram tiling and searching. During each decoding time step the decoder produces a softmax distribution over the input sequence (see Figure \ref{pointer_network}), which in our implementation has a maximum sequence length of 3000. Additionally the \texttt{BEGIN, END, PAD} symbols add to a total of \texttt{3003} symbols to softmax over. The cross entropy loss function is averaged over the entire output sequence of labels and is considered the final loss for the entire input sequence.

\subsection {Network Configuration}

We use a single layer bi-LSTM~\cite{10.5555/1986079.1986220} pointer network with 512 hidden units in a layer and an attention size of 128. Addition of an extra layer to the network did not result in an improvement. The Adam optimizer \cite{kingma2014adam} was used with an initial learning rate of 0.001. A maximum input sequence length of 3000 and a maximum output length of 100 were enforced.

\section{Datasets}

For reasons explained in section 1 we exclusively evaluate on Wikidata based datasets. We use the following:




\begin{itemize}

\item\textbf{WebQSP}: We use the dataset released by Sorokin et al \cite{sorokin-gurevych-2018-mixing} where the original WebQSP dataset by Yih et al \cite{yih-etal-2016-value}, which was based on Freebase, has been adapted and all Freebase IDs converted to their respective Wikidata IDs. WebQSP contains questions that were originally collected for the WebQuestions dataset from web search logs (Berant et al \cite{berant-etal-2013-semantic}). WebQSP is a relatively small dataset consisting of 3098 train 1639 test questions which cover 3794 and 2002 entities respectively. The dataset has a mixture of simple and complex questions. We found some questions in the test set that had failed Freebase to Wikidata entity ID conversions. We skipped such questions during PNEL's evaluation.


\item\textbf{SimpleQuestions}: To test the performance of PNEL on simple questions, we choose SimpleQuestions \cite{bordes2015largescale}, which as the name suggests, consists only of Simple Questions. The training set has more than 30,000 questions while the test set has close to 10,000 questions. This dataset was also originally based on Freeebase and later the entity IDs were converted to corresponding Wikidata IDs. However out of the 10,000 test questions only about half are answerable on the current Wikidata. 


\item\textbf{LC-QuAD 2.0}: Unlike the first two datasets, LC-QuAD 2.0 \cite{10.1007/978-3-030-30796-7_5} is based on Wikidata since its inception and is also the most recent dataset of the three. It carries a mixture of simple and complex questions which were verbalised by human workers on Amazon Mechanical Turk. It is a large and varied dataset comprising of 24180 train questions and 6046 test questions which cover 33609 and 8417 entities respectively.

\end{itemize}

\section{Evaluation}


In this section we evaluate our proposed model(s) against different state-of-the-art methods for KGQA. As notations, PNEL-L stands for PNEL trained on LC-QuAD 2.0. PNEL-W and PNEL-S stand for PNEL trained on WebQSP and SimpleQuestions respectively.\\


\subsection{Experiment 1 : EL over KBPearl split of LC-QuAD 2.0 test set}
\textbf{Objective:} The purpose of this experiment is to benchmark PNEL against a large number of EL systems not just over Wikidata but also other KBs.\\
\textbf{Method:} The results are largely taken from KBPearl. PNEL is trained on the LC-QuAD 2.0 training set. For a fair comparison, the systems are tested on the 1294 questions split of test set provided by KBPearl. We train PNEL for 2 epochs.\\
\textbf{Remarks:} Results for Falcon 2.0 and OpenTapioca were obtained by accessing their live API. The original Falcon 2.0 paper provides an F1 of 0.69 on 15\% of randomly selected questions from a combination of the train and test splits of the dataset. Several systems in the table below do not originally produce Wikidata entity IDs, hence the authors of KBpearl have converted the IDs to corresponding Wikidata IDs.
\\
\textbf{Analysis: } As observed from the results in Table \ref{lcqeval1}, PNEL outperforms all other systems on this particular split of LC-QuAD 2.0 dataset.

\begin{table} [h]
\centering 
\begin{tabular}{ p{3cm} p{1.9cm} p{1.9cm} p{1.5cm}  }
\toprule
 Entity Linker & Precision & Recall &F1 \\
 \toprule
 Falcon\cite{sakor-etal-2019-old}        &0.533 &0.598 &0.564   \\
 EARL\cite{10.1007/978-3-030-00671-6_7}          &0.403 &0.498 &0.445 \\
 Spotlight\cite{DBLP:conf/i-semantics/MendesJGB11}     &0.585 &0.657 &0.619 \\
 TagMe\cite{ferragina2010tagme}         &0.352 &\textbf{0.864} &0.500 \\
 OpenTapioca\cite{delpeuch2019opentapioca}   &0.237 &0.411 &0.301 \\
 QKBfly\cite{nguyen2017query}        &0.518 &0.479 &0.498 \\ 
 Falcon 2.0    &0.395 &0.268 &0.320 \\
 KBPearl-NN    &0.561 &0.647 &0.601 \\
 \midrule
 PNEL-L    &\textbf{0.803} &0.517 &\textbf{0.629}  \\
\bottomrule
\end{tabular}\\
\captionof{table}{Evaluation on KBPearl split of LC-QuAD 2.0 test set}
\label{lcqeval1}
\end{table}

\subsection{Experiment 2 : EL over full LC-QuAD 2.0 test set}
\textbf{Objective:} The objective of this experiment is to compare systems that return Wikidata IDs for the EL task.\\
\textbf{Method:}  We train PNEL on LC-QuAD 2.0 train set and test on all 6046 questions in test set. PNEL was trained for 2 epochs. \\
\textbf{Remarks:} Results for competing systems were obtained by accessing their live APIs. We choose systems that return Wikidata IDs. \\
\textbf{Analysis: } As seen in Table \ref{lcqeval2}, similar to the previous experiment, PNEL performs the best on the LC-QuAD 2.0 test set.

\begin{table}[!h]
\centering 
     \begin{tabular}{ p{2.7cm} p{2.1cm} p{2.1cm} p{1.8cm}   }
 \toprule
 Entity Linker &Precision &Recall &F1 \\
 \toprule
 VCG\cite{sorokin-gurevych-2018-mixing} &0.516 &0.432 &0.470 \\ 
 OpenTapioca\cite{delpeuch2019opentapioca}   &0.237  & 0.411 &0.301  \\
 Falcon 2.0    &0.418 &0.476 &0.445 \\
 \midrule
 PNEL-L    &\textbf{0.688} &\textbf{0.516} &\textbf{0.589} \\
 \bottomrule
\end{tabular}\\
\caption{Evaluation on LC-QuAD 2.0 test set}
\label{lcqeval2}
\end{table}

\subsection{Experiment 3 : EL over WebQSP test set}

\textbf{Objective:} Benchmark against an end-to-end model that returns Wikidata IDs.\\
\textbf{Method:} Train and test PNEL on WebQSP train and test sets respectively. PNEL is trained for 10 epochs.\\
\textbf{Remarks:} Results for the competing systems were taken from Sorokin et al \cite{sorokin-gurevych-2018-mixing}.\\

\begin{table} [h]
\centering 
\begin{tabular}{ p{2.8cm} p{1.8cm} p{1.8cm} p{1.8cm}   }
 
\toprule
 Entity Linker &Precision &Recall &F1 \\
\toprule
 Spotlight     &0.704 &0.514 &0.595   \\
 S-MART\cite{yang2016s}        &0.666 &\textbf{0.772} &0.715  \\
 VCG\cite{sorokin-gurevych-2018-mixing}           &0.826 &0.653 &\textbf{0.730}  \\
 \midrule
 
 PNEL-L         &0.636 &0.480 &0.547   \\ 
 PNEL-W     &\textbf{0.886} &0.596 &0.712 \\
 \bottomrule
\end{tabular}
\caption {Evaluation on WebQSP}
\label{webqeval}
\end{table}

\textbf{Analysis: } As seen in Table \ref{webqeval} PNEL comes in third best in this experiment, beaten by VCG and S-MART. S-MART has high recall because it performs semantic information retrieval apart from lexical matching for candidate generation, as explained in Section~\ref{relatedwork}. VCG is more similar to PNEL in that it is also an end-to-end system. It has higher recall but lower precision than PNEL. 



\subsection{Experiment 4 : EL over SimpleQuestions test set}

\textbf{Objective:} Benchmark systems on the SimpleQuestions Dataset. \\
\textbf{Method:} Train and test PNEL on SimpleQuestions train and test sets respectively. PNEL is trained for 2 epochs.\\
\textbf{Remarks: }  We extended the results from Falcon 2.0~\cite{sakor2019falcon}. \\

\begin{table} [h]
\centering 
\begin{tabular}{ p{3cm} p{1.8cm} p{1.8cm} p{1.8cm}  }
 \toprule
 Entity Linker &Precision &Recall &F1  \\
 \toprule
 OpenTapioca\cite{delpeuch2019opentapioca}     &0.16 &0.28 &0.20  \\
 Falcon 2.0      &0.38 &0.44 &0.41  \\
 \midrule
 PNEL-L            &0.31 &0.25 &0.28  \\
 PNEL-S           &\textbf{0.74} &\textbf{0.63} &\textbf{0.68} \\
 \bottomrule
\end{tabular}
\caption{Evaluation on SimpleQuestions}
\label{simpleqeval}
\end{table}


\textbf{Analysis: } As seen in Table \ref{simpleqeval}, PNEL outperforms the competing systems both in precision and recall for SimpleQuestions dataset. As observed, PNEL has the best precision across all datasets, however, recall seems to be PNEL's weakness.

\subsection{Experiment 5 : Candidate generation accuracy}

\textbf{Objective:} The purpose of this experiment is to see what percentage of correct entity candidates were made available to PNEL after the text search phase. This sets a limit on the maximum performance that can be expected from PNEL.\\ 
\textbf{Remarks:} PNEL considers each token a possible correct entity, but since it only considers top-K text search matches for each token, it also loses potentially correct entity candidates before the disambiguation phase. The results in Table \ref{spanloss} are for K=30. \\

\begin{table} [h]
\centering 
\begin{tabular}{  p{2.7cm} p{1.9cm}  }
 \toprule
 Dataset & PNEL (\%)  \\
 \toprule
 WebQSP      &73 \\
 LC-QuAD 2.0         &82 \\
 SimpleQuestions   &90 \\
 \bottomrule
\end{tabular}
\caption{Entity Candidates available post label search}
\label{spanloss}
\end{table}

\subsection{Experiment 6 : Ablation of features affecting accuracy}

\textbf{Objective:} We present an ablation study on the WebQSP dataset to understand the importance of different feature vectors used in the model.


\begin{table} [h]
\centering 
\begin{tabular}{ p{1.6cm} p{1.2cm} p{1.6cm} p{1.2cm} p{1.0cm} p{1.0cm} p{1.3cm} p{1.3cm} p{1.1cm}  }
 \toprule
 Sentence Embed. &Word Embed. &Descript. Embed. &TransE  &PoS Tags & Text Rank &n-gram length&Text Match Metric&F1 Score \\
 \toprule
   \checkmark & \checkmark   &\checkmark &\checkmark &\checkmark & \checkmark & \checkmark& \checkmark &0.712 \\
    & \checkmark   &\checkmark &\checkmark &\checkmark & \checkmark & \checkmark&\checkmark&0.554\\
   \checkmark &    &\checkmark &\checkmark &\checkmark & \checkmark & \checkmark&\checkmark&0.666\\
   \checkmark & \checkmark   & &\checkmark &\checkmark & \checkmark & \checkmark&\checkmark&0.700 \\
   \checkmark & \checkmark   &\checkmark & &\checkmark & \checkmark & \checkmark&\checkmark&\textbf{0.221} \\
   \checkmark & \checkmark   &\checkmark &\checkmark & & \checkmark & \checkmark&\checkmark&0.685 \\
   \checkmark & \checkmark   &\checkmark &\checkmark &\checkmark &  & \checkmark&\checkmark&0.399 \\
   \checkmark & \checkmark   &\checkmark &\checkmark &\checkmark & \checkmark & &\checkmark&0.554 \\
   \checkmark & \checkmark   &\checkmark &\checkmark &\checkmark & \checkmark & \checkmark&  &0.698 \\
 \bottomrule
\end{tabular}
\caption{Ablation test for PNEL on WebQSP test set features}
\label{pnelablation}
\end{table}

\textbf{Analysis:} As seen in Table \ref{pnelablation} it appears that the most important feature is the TransE entity embedding, which implicitly contains the entire KG structure information. On removing this feature there is drop in F1 score from 0.712 to 0.221. On the other hand the least important feature seem to be the description embedding. Removal of this feature merely leads to a drop in F1 from 0.712 to 0.700. A possible reason is that the Text Search Rank potentially encodes 
significant text similarity information, and TransE potentially encodes other type and category related information that description often adds. Removal of the Text Search Rank also results in a large drop in F1 reaching to 0.399 from 0.712.

\subsection{Experiment 7 : Run Time Evaluation}

\textbf{Objective:} We look at a comparison of run times across the systems we have evaluated on


\begin{table} [h]
\centering 
\begin{tabular}{ p{2.5cm} p{2cm} p{2.6cm} p{4cm}  }
    \toprule
    System &Seconds & Target KG \\
    \toprule
        VCG & 8.62  & Wikidata    \\
        PNEL & 3.14 & Wikidata   \\
    \midrule
        Falcon 2.0 & 1.08 & Wikidata    \\
        EARL & 0.79 & DBpedia   \\
        TagME & 0.29 & Wikipedia   \\
        Spotlight & 0.16 & DBpedia    \\
        Falcon & 0.16 & DBpedia    \\
        OpenTapioca & 0.07 & Wikidata    \\
 \bottomrule
\end{tabular}
\caption{Time taken per question on the WebQSP dataset of 1639 questions}
\label{pnelruntime}
\end{table}

\vspace{0.1mm}

\textbf{Analysis:} QKBFly and KBPearl are off-line systems, requiring separate steps for entity candidate population and entity linking, hence they are not evaluated in Table \ref{pnelruntime}. VCG and PNEL are end-to-end systems while the others are modular systems.  VCG and PNEL were installed locally on a machine with the following configuration: 256 GB RAM, 42 core E5-2650 Intel Xeon v4@2.2GHz. No GPU was present on the system during run time. For VCG and PNEL, the times taken for first runs were recorded, where the corresponding databases such as Virtuoso and Elasticsearch, were started just before the evaluation. This was done so that the times were not affected by caching from previous runs. For systems except PNEL and VCG, the times mentioned in the table were collected from API calls to their hosted services. It must be considered that, due to network latency, and other unknown setup related configurations at the service end, the times may not be comparably directly. PNEL performs faster than VCG since it avoids querying the KG during runtime, and instead relies on pre-computed KG embeddings. PNEL also uses lesser number of features than VCG. A visible trend is that the more accurate system is slower, however Spotlight is an exception, which performs well in both speed and accuracy.

\section{Error Analysis}

A prominent feature of PNEL is high precision and low recall. We focus on loss in recall in this section. For LC-QuAD 2.0 test set consisting of 6046 questions, the precision, recall and F-score are 0.688, 0.516 and 0.589 respectively. We categorise the phases of loss in recall in two sections  1) Failure in the candidate generation phase 2) Failure in re-ranking/disambiguation phase. When considering the top 50 search candidates during text label search, it was found that 75.3\% of the correct entities were recovered from the entity label index. This meant that before re-ranking we had already lost 24.7\% recall accuracy. During re-ranking phase, a further 23.7\% in absolute accuracy was lost, leading to our recall of 0.516. We drill down into the 23.7\% absolute loss in accuracy during re-ranking, attempting to find the reasons for such loss, since this would expose the weaknesses of the model. In the plots below, we consider all those questions which contained the right candidate entity in the candidate generation phase. Hence, we discard those questions for our analysis, which already failed in the candidate generation phase.

\begin{table} [h]
\centering 
\begin{tabular}{ p{2.3cm} p{2.5cm} p{2cm} p{2cm} p{2cm}  }
    \toprule
    Entity Count & Questions Count  &Precision  &Recall &F1 \\
    \toprule
        1 & 3311 &0.687& 0.636& 0.656  \\
        2 & 1981& 0.774& 0.498 & 0.602 \\
        3 & 88& 0.666& 0.431& 0.518  \\
 \bottomrule
\end{tabular}
\caption{Comparison of PNEL's performance with respect to number of entities in a question.}

\label{recallsimplecomplex}
\vspace{-10pt}
\end{table}

\vspace{0.1mm}

\begin{figure*}[thb]
  \centering
  \begin{minipage}[b]{0.48\textwidth}
    \includegraphics[width=1.0\textwidth]{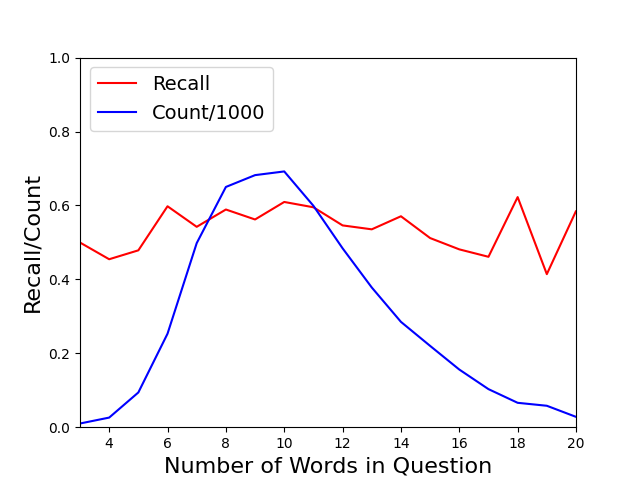}
  \end{minipage}
  \begin{minipage}[b]{0.48\textwidth}
    \includegraphics[width=1.0\textwidth]{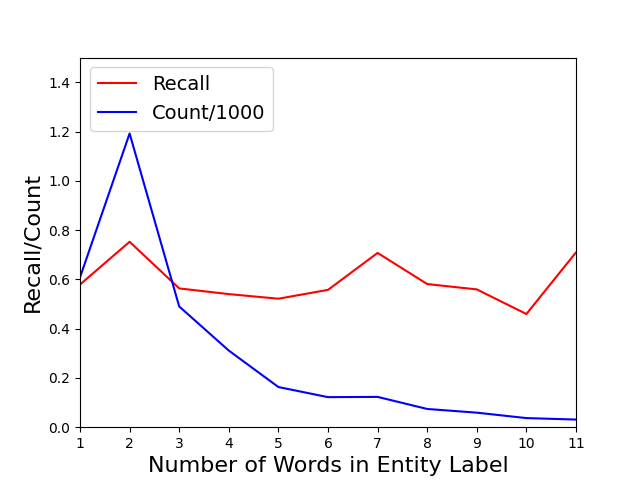}
  \end{minipage}
  
      \caption{Plots of recall variation versus 1) Length of Question 2) Length of entity span 3) Frequency of questions with the given lengths in the dataset (scaled down by a factor of 1000)}
\label{recallplots}
\vspace{-10pt}
\end{figure*}

It is observed in Table \ref{recallsimplecomplex} that recall falls as the number of entities per question rises. It must not be concluded however, that PNEL fails to recognise more than an entity per question. There were 375 questions with multiple entities where PNEL was able to link all the entities correctly. In Figure \ref{recallplots} we observe that the recall does not exhibit significant co-relation with either the length of the question, or the length of entity label. The recall remains stable. There seems to be some co-relation between the amount of data available for a given length of question, and the recall on it. It appears that the model performs better on question lengths it has seen more often during training.

\section{Discussion and Future Work}

In this work we have proposed PNEL, an end-to-end Entity Linking system based on the Pointer Network model. We make no modifications to the original Pointer Network model, but identify its utility for the problem statement of EL, and successfully model the problem so the Pointer Network is able to find the right set of entities.   We evaluate our approach on three datasets of varying complexity and report state of the art results on two of them. On the third dataset, WebQSP, we perform best in precision but lag behind in recall. We select such features that require no real time KG queries during inference. This demonstrates that the Pointer Network model, and the choice of features presented in this work, result in a practical and deployable EL solution for the largest Knowledge Graph publicly available - Wikidata.   \\


For future work: PNEL being based on the LSTM cell inevitably processes tokens sequentially increasing the response times. This limitation could be overcome by using some variant of the Transformer model~\cite{NIPS2017_7181} instead, which is not only a powerful model but also able to process tokens in parallel.  As a future work we would also like to explore different entity embedding techniques and investigate which characteristics make an embedding suitable for the entity linking task. 
\vspace{-10pt}
\section*{Acknowledgement}
We would like to thank Prof. Dr. Chris Biemann of the Language Technology Group, University of Hamburg, for his valuable suggestions towards improving this work.
\vspace{-10pt}

\bibliographystyle{abbrv}
\bibliography{pnel}

\begin{thebibliography}{10}

\bibitem{berant-etal-2013-semantic}
J.~Berant, A.~Chou, R.~Frostig, and P.~Liang.
\newblock {Freebase from Question-Answer Pairs}.
\newblock In {\em Proceedings of the 2013 Conference on Empirical Methods in
  Natural Language Processing}. Association for Computational Linguistics,
  2013.

\bibitem{bollacker2008freebase}
K.~Bollacker, C.~Evans, P.~Paritosh, T.~Sturge, and J.~Taylor.
\newblock {Freebase: A Collaboratively Created Graph Database for Structuring
  Human Knowledge}.
\newblock In {\em Proceedings of the 2008 ACM SIGMOD international conference
  on Management of data}. AcM, 2008.

\bibitem{bordes2015largescale}
A.~Bordes, N.~Usunier, S.~Chopra, and J.~Weston.
\newblock {Large-scale Simple Question Answering with Memory Networks}, 2015.

\bibitem{delpeuch2019opentapioca}
A.~Delpeuch.
\newblock {OpenTapioca: Lightweight Entity Linking for Wikidata}.
\newblock {\em arXiv preprint arXiv:1904.09131}, 2019.

\bibitem{10.1007/978-3-030-30796-7_5}
M.~Dubey, D.~Banerjee, A.~Abdelkawi, and J.~Lehmann.
\newblock Lc-quad 2.0: A large dataset for complex question answering over
  wikidata and dbpedia.
\newblock In {\em The Semantic Web -- ISWC 2019}. Springer International
  Publishing, 2019.

\bibitem{10.1007/978-3-030-00671-6_7}
M.~Dubey, D.~Banerjee, D.~Chaudhuri, and J.~Lehmann.
\newblock {EARL: Joint Entity and Relation Linking for Question Answering over
  Knowledge Graphs}.
\newblock In {\em The Semantic Web -- ISWC 2018}. Springer International
  Publishing, 2018.

\bibitem{10.1145/1871437.1871689}
P.~Ferragina and U.~Scaiella.
\newblock {TAGME: On-the-Fly Annotation of Short Text Fragments (by Wikipedia
  Entities)}.
\newblock 2010.

\bibitem{ferragina2010tagme}
P.~Ferragina and U.~Scaiella.
\newblock {Tagme: On-the-Fly Annotation of Short Text Fragments (by Wikipedia
  Entities)}.
\newblock In {\em Proceedings of the 19th ACM international conference on
  Information and knowledge management}, 2010.

\bibitem{10.1007/978-3-030-30796-7_7}
J.~Frey, M.~Hofer, D.~Obraczka, J.~Lehmann, and S.~Hellmann.
\newblock {DBpedia FlexiFusion the Best of Wikipedia > Wikidata > Your Data}.
\newblock In {\em The Semantic Web -- ISWC 2019}. Springer International
  Publishing, 2019.

\bibitem{10.5555/1986079.1986220}
A.~Graves, S.~Fern\'{a}ndez, and J.~Schmidhuber.
\newblock {Bidirectional LSTM Networks for Improved Phoneme Classification and
  Recognition}.
\newblock In {\em Proceedings of the 15th International Conference on
  Artificial Neural Networks: Formal Models and Their Applications - Volume
  Part II}. Springer-Verlag, 2005.

\bibitem{kingma2014adam}
D.~P. Kingma and J.~Ba.
\newblock {Adam: A Method for Stochastic Optimization}.
\newblock {\em arXiv preprint arXiv:1412.6980}, 2014.

\bibitem{Kolitsas_2018}
N.~Kolitsas, O.-E. Ganea, and T.~Hofmann.
\newblock {End-to-End Neural Entity Linking}.
\newblock {\em Proceedings of the 22nd Conference on Computational Natural
  Language Learning}, 2018.

\bibitem{lehmann2015dbpedia}
J.~Lehmann, R.~Isele, M.~Jakob, A.~Jentzsch, D.~Kontokostas, P.~N. Mendes,
  S.~Hellmann, M.~Morsey, P.~Van~Kleef, S.~Auer, et~al.
\newblock {DBpedia -- A Large-Scale, Multilingual Knowledge Base Extracted from
  Wikipedia}.
\newblock {\em Semantic Web}, 2015.

\bibitem{pbg}
A.~Lerer, L.~Wu, J.~Shen, T.~Lacroix, L.~Wehrstedt, A.~Bose, and
  A.~Peysakhovich.
\newblock {PyTorch-BigGraph: A Large-scale Graph Embedding System}.
\newblock In {\em Proceedings of the 2nd SysML Conference}, 2019.

\bibitem{10.14778/3384345.3384352}
X.~Lin, H.~Li, H.~Xin, Z.~Li, and L.~Chen.
\newblock {KBPearl: A Knowledge Base Population System Supported by Joint
  Entity and Relation Linking}.
\newblock {\em Proc. VLDB Endow.}, 2020.

\bibitem{luo-etal-2015-joint}
G.~Luo, X.~Huang, C.-Y. Lin, and Z.~Nie.
\newblock {Joint Entity Recognition and Disambiguation}.
\newblock In {\em Proceedings of the 2015 Conference on Empirical Methods in
  Natural Language Processing}. Association for Computational Linguistics,
  2015.

\bibitem{DBLP:conf/i-semantics/MendesJGB11}
P.~N. Mendes, M.~Jakob, and Andr{\'{e}s Garc{\'{\i}a{-}Silva and Christian
  Bizer}}.
\newblock {DBpedia Spotlight: Shedding Light on the Web of Documents}.
\newblock In {\em Proceedings the 7th International Conference on Semantic
  Systems}, 2011.

\bibitem{nguyen2017query}
D.~B. Nguyen, A.~Abujabal, K.~Tran, M.~Theobald, and G.~Weikum.
\newblock {Query-Driven On-the-Fly Knowledge Base Construction}.
\newblock {\em Proceedings of the VLDB Endowment}, 2017.

\bibitem{Nguyen:2017:QOK:3151113.3151119}
D.~B. Nguyen, A.~Abujabal, N.~K. Tran, M.~Theobald, and G.~Weikum.
\newblock {Query-driven On-the-fly Knowledge Base Construction}.
\newblock {\em Proc. VLDB Endow.}, 2017.

\bibitem{nguyen-etal-2016-j}
D.~B. Nguyen, M.~Theobald, and G.~Weikum.
\newblock {NERD: Joint Named Entity Recognition and Disambiguation with Rich
  Linguistic Features}.
\newblock {\em Transactions of the Association for Computational Linguistics},
  2016.

\bibitem{10.1145/2872427.2874809}
T.~Pellissier~Tanon, D.~Vrande\v{c}i\'{c}, S.~Schaffert, T.~Steiner, and
  L.~Pintscher.
\newblock {From Freebase to Wikidata: The Great Migration}.
\newblock In {\em Proceedings of the 25th International Conference on World
  Wide Web}. International World Wide Web Conferences Steering Committee, 2016.

\bibitem{Perozzi_2014}
B.~Perozzi, R.~Al-Rfou, and S.~Skiena.
\newblock {DeepWalk}.
\newblock {\em Proceedings of the 20th ACM SIGKDD international conference on
  Knowledge discovery and data mining - KDD ’14}, 2014.

\bibitem{sakor-etal-2019-old}
A.~Sakor, I.~Onando~Mulang{'}, K.~Singh, S.~Shekarpour, M.~Esther~Vidal,
  J.~Lehmann, and S.~Auer.
\newblock Old is gold: Linguistic driven approach for entity and relation
  linking of short text.
\newblock In {\em Proceedings of the 2019 Conference of the North {A}merican
  Chapter of the Association for Computational Linguistics: Human Language
  Technologies, Volume 1 (Long and Short Papers)}. Association for
  Computational Linguistics, 2019.

\bibitem{sakor2019falcon}
A.~Sakor, K.~Singh, A.~Patel, and M.-E. Vidal.
\newblock {FALCON 2.0: An Entity and Relation Linking Tool over Wikidata},
  2019.

\bibitem{santorini1990part}
B.~Santorini.
\newblock {Part-of-speech tagging guidelines for the Penn Treebank Project}.
\newblock 1990.

\bibitem{sevgili-etal-2019-improving}
{\"O}.~Sevgili, A.~Panchenko, and C.~Biemann.
\newblock {Improving Neural Entity Disambiguation with Graph Embeddings}.
\newblock In {\em Proceedings of the 57th Annual Meeting of the Association for
  Computational Linguistics: Student Research Workshop}. Association for
  Computational Linguistics, 2019.

\bibitem{10.1145/2505515.2505601}
A.~Sil and A.~Yates.
\newblock {Re-Ranking for Joint Named-Entity Recognition and Linking}.
\newblock In {\em Proceedings of the 22nd ACM International Conference on
  Information and Knowledge Management}. Association for Computing Machinery,
  2013.

\bibitem{sorokin-gurevych-2018-mixing}
D.~Sorokin and I.~Gurevych.
\newblock {Mixing Context Granularities for Improved Entity Linking on Question
  Answering Data across Entity Categories}.
\newblock In {\em Proceedings of the Seventh Joint Conference on Lexical and
  Computational Semantics}. Association for Computational Linguistics, 2018.

\bibitem{suchanek2007semantic}
F.~M. Suchanek, G.~Kasneci, and G.~Weikum.
\newblock {Yago: A Core of Semantic Knowledge}.
\newblock In {\em Proceedings of the 16th International Conference on World
  Wide Web}. ACM, 2007.

\bibitem{NIPS2017_7181}
A.~Vaswani, N.~Shazeer, N.~Parmar, J.~Uszkoreit, L.~Jones, A.~N. Gomez, L.~u.
  Kaiser, and I.~Polosukhin.
\newblock Attention is all you need.
\newblock In I.~Guyon, U.~V. Luxburg, S.~Bengio, H.~Wallach, R.~Fergus,
  S.~Vishwanathan, and R.~Garnett, editors, {\em Advances in Neural Information
  Processing Systems 30}, pages 5998--6008. Curran Associates, Inc., 2017.

\bibitem{NIPS2015_5866}
O.~Vinyals, M.~Fortunato, and N.~Jaitly.
\newblock Pointer networks.
\newblock In C.~Cortes, N.~D. Lawrence, D.~D. Lee, M.~Sugiyama, and R.~Garnett,
  editors, {\em Advances in Neural Information Processing Systems 28}. Curran
  Associates, Inc., 2015.

\bibitem{vrandevcic2014wikidata}
Vrande{\v{c}i{\'c}, Denny and Kr{\"o}tzsch, Markus}.
\newblock {Wikidata: A Free Collaborative Knowledge Base}.
\newblock {\em Communications of the ACM}, 2014.

\bibitem{smart2015Yang}
Y.~Yang and M.-W. Chang.
\newblock {S-MART: Novel Tree-based Structured Learning Algorithms Applied to
  Tweet Entity Linking.}
\newblock In {\em ACL 2015}, 2015.

\bibitem{yang2016s}
Y.~Yang and M.-W. Chang.
\newblock {S-Mart: Novel Tree-Based Structured Learning Algorithms Applied to
  Tweet Entity Linking}.
\newblock {\em arXiv preprint arXiv:1609.08075}, 2016.

\bibitem{yih-etal-2016-value}
W.-t. Yih, M.~Richardson, C.~Meek, M.-W. Chang, and J.~Suh.
\newblock The value of semantic parse labeling for knowledge base question
  answering.
\newblock In {\em Proceedings of the 54th Annual Meeting of the Association for
  Computational Linguistics (Volume 2: Short Papers)}. Association for
  Computational Linguistics, 2016.

\end{thebibliography}
\end{document}